\documentclass{article}
\usepackage{spconf,amsmath,graphicx,hyperref}
\usepackage{amssymb}
\usepackage{xcolor}
\usepackage{booktabs}
\usepackage{subcaption}
\usepackage{adjustbox}
\usepackage{caption}
\usepackage{multirow}

\DeclareMathOperator{\softmax}{softmax}

\title{EMoE: Eigenbasis-Guided Routing for Mixture-of-Experts}
\name{
\begin{tabular}[t]{c}
    Anzhe Cheng\quad
    Shukai Duan\quad
    Shixuan Li\quad
    Chenzhong Yin\quad \\
    Mingxi Cheng\quad
    Shahin Nazarian\quad
    Paul Thompson\quad
    Paul Bogdan
       \end{tabular}}
\address{University of Southern California, Los Angeles, CA, USA}
\begin{document}
%
\maketitle
\begin{abstract}
\vspace{-3mm}
The relentless scaling of deep learning models has led to unsustainable computational demands, positioning Mixture-of-Experts (MoE) architectures as a promising path towards greater efficiency. However, MoE models are plagued by two fundamental challenges: 1) a load imbalance problem known as the``rich get richer" phenomenon, where a few experts are over-utilized, and 2) an expert homogeneity problem, where experts learn redundant representations, negating their purpose. Current solutions typically employ an auxiliary load-balancing loss that, while mitigating imbalance, often exacerbates homogeneity by enforcing uniform routing at the expense of specialization. To resolve this, we introduce the Eigen-Mixture-of-Experts (EMoE), a novel architecture that leverages a routing mechanism based on a learned orthonormal eigenbasis. EMoE projects input tokens onto this shared eigenbasis and routes them based on their alignment with the principal components of the feature space. This principled, geometric partitioning of data intrinsically promotes both balanced expert utilization and the development of diverse, specialized experts, all without the need for a conflicting auxiliary loss function. Our code is publicly available at \url{https://github.com/Belis0811/EMoE}.
\end{abstract}
\vspace{-4mm}
\section{Introduction}
\label{sec:intro}
\vspace{-3mm}
Deep learning has emerged as a cornerstone of modern artificial intelligence, driving state-of-the-art advances in computer vision and beyond~\cite{chai2021deep, lecun2015deep, hinton2012deep}. This success is largely attributed to a straightforward scaling principle: training larger models on more data consistently leads to improved performance~\cite{kaplan2020scaling}. However, this paradigm has led to an exponential and unsustainable growth in model size and computational cost. The compute required to train frontier AI models has been doubling approximately every six months since 2010, far outpacing improvements in hardware and incurring prohibitive financial and environmental costs~\cite{sevilla2022compute, powerai, schwartz2020green}. One promising solution to this scaling challenge is the Mixture-of-Experts (MoE) architecture~\cite{shazeer2017outrageously}, which uses conditional computation to activate only a small subset of specialized "expert" sub-networks for each input. By doing so, MoE models can dramatically increase their total parameter count while keeping the inference cost (FLOPs) nearly constant, effectively decoupling model capacity from runtime cost.

\begin{figure*}[htbp]
 \centering
    \includegraphics[width=\textwidth]{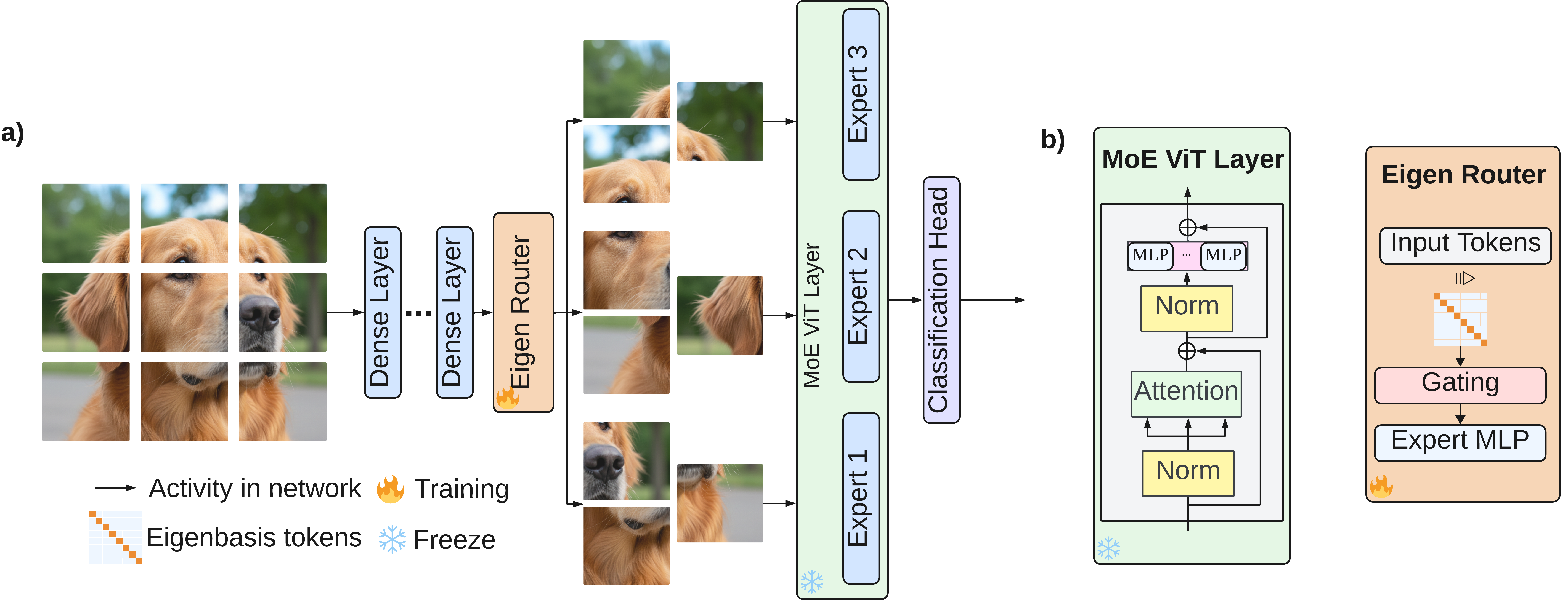}
    \vspace{-6mm}
\caption{\textbf{Overview of EMoE.} (a) Training schematic: the eigenbasis is recomputed from routers with frozen ViT weight (b) EMoE-augmented ViT block: input tokens convert into eigenbasis tokens, then pass through LayerNorm and multi-head self-attention, followed by a MoE feed-forward layer.}
\label{fig:eigenbasis_routing}
\vspace{-5mm}
\end{figure*}

Despite their potential, MoE models often suffer from degenerative behaviors that limit their practical effectiveness~\cite{fedus2022switch, riquelme2021scaling}. The most prominent issue is routing collapse – a``rich-get-richer" effect where the gating mechanism routes a disproportionate number of inputs to a few popular experts~\cite{zhou2022mixture, liu2024routers}. This severe load imbalance creates computational bottlenecks: throughput is dictated by the busiest expert, leaving much of the model's capacity underutilized. The common remedy is to add an auxiliary Load-Balancing Loss (LBL)\cite{wang2024auxiliary} to the training objective, encouraging a more uniform distribution of inputs across experts. However, recent analyses have shown that enforcing uniform routing in this way can conflict with the goal of expert specialization\cite{guo2025advancing, qiu2025demons}. By forcing the router to prioritize even load over content, LBL may cause experts to learn redundant, homogeneous representations (with pairwise similarities reported as high as 99\%). This undermines the intended divide-and-conquer principle of the MoE approach. In essence, current MoE training techniques face a trade-off: they achieve either balanced loads or highly specialized experts, but not both simultaneously.

To resolve this tension between computational balance and representational diversity, we propose the Eigenbasis-Guided Routing for Mixture-of-Experts (EMoE). EMoE fundamentally reimagines the routing mechanism by eliminating the conventional learned gating network and its auxiliary loss. Instead, routing decisions are made by projecting input features onto an eigenbasis derived from the data (the principal components of the feature distribution). This geometric partitioning of the feature space naturally assigns coherent clusters of inputs to different experts. By design, our approach promotes both balanced expert utilization and specialized expert representations without the need for a conflicting auxiliary objective. We evaluate EMoE on the ImageNet image classification benchmark, comparing it against state-of-the-art MoE models including V-MoE~\cite{riquelme2021scaling}, a single-gated MoE variant~\cite{royer2023revisiting}, and DeepMoE~\cite{wang2020deep}. EMoE attains accuracy on par with these models while achieving significantly better load balance across experts, demonstrating that balanced routing and expert specialization can be achieved simultaneously for large-scale image classification.

\vspace{-3mm}
\section{Method}
\label{sec:method}
\vspace{-3mm}
\textbf{Overview of the architecture.} Our proposed method integrates a sparse MoE layer into the ViT backbone with eigenbasis guide routers. Each MoE branch consists of an Eigen Router and a set of 8 expert MLP networks. The branch operates in parallel with the ViT’s feed-forward layer; the patch token outputs are fed into the Eigen Router, which dynamically selects one expert MLP per token. The chosen expert transforms the token’s features, and the expert’s output is added back to the token’s representation. In practice, the router selects top expert per token, sends the token through that expert’s MLP, optionally scales by the gate score, and then adds the result back through the residual stream. Importantly, only the patch tokens pass through the MoE branch, i.e., the class token bypasses the router and continues along the standard transformer pathway unchanged. Fig.~\ref{fig:eigenbasis_routing}a illustrates this architecture, where each ViT block in the augmented set has a parallel EMoE route comprising the eigenbasis-guided router and multiple expert MLPs.

\textbf{Eigen Router.} The Eigen Router is a novel gating mechanism that routes tokens to experts based on the covariance structure of their features. As shown in Fig.~\ref{fig:eigenbasis_routing}b, at each forward pass, the router collects the patch token features output by the ViT block. Conceptually, given a batch of images, let $H \in \mathbb{R}^{N\times D}$ be the matrix of patch token features(with $N$ tokens across the batch and $D$ dimensional embeddings per token), and the router maintains a learnable eigenbasis matrix $\mathbf{U}\in\mathbb{R}^{D\times r}$ ($r \ll D$) that is optimized to align with the dominant eigenspace of the empirical covariance $\mathbf{C} = \frac{1}{N}H^\top H$. In other words, $\mathbf{U}$ spans the top-$r$ principal components of the patch tokens’ feature distribution. We enforce this through an orthonormality regularizer that keeps $\mathbf{U}$ close to orthonormal, which encourages $\mathbf{U}^\top \mathbf{U} \approx \mathbf{I}_r$. This learned eigenbasis is updated continuously by reorthognal the input tokens during each step of training, allowing the router to adapt to the feature statistics of each batch.

Given the basis $\mathbf{U}$, the router projects each patch token’s feature vector $h_t \in \mathbb{R}^D$ into the $r$-dimensional subspace. Let $z_t = h_t^\top \mathbf{U} \in \mathbb{R}^r$ denote the coordinates of token $t$ in the eigenbasis. We compute the normalized energy of $h_t$ along each basis direction $j$ as the squared projection, normalized to form a distribution:
\begin{equation}
    e_{t,j} = \frac{z^2_{t,j}}{\sum_{k=1}^r z^2_{t,k} + \epsilon},
\end{equation}

where $\epsilon$ is a small constant for numerical stability. The vector $e_t = (e_{t,1},\dots,e_{t,r})$ of dimension $r$ represents the fraction of the token’s feature energy captured by each eigenvector. Intuitively, $e_t$ tells us how “aligned” the token is with each principal direction of variance.

Next, the router transforms these per-token basis-coefficient vectors into expert logits for gating. We associate each expert $k$ (for $k=1,\dots,8$) with a learned weight vector over the $r$ dimensions. In implementation, this is done via a learned weight matrix $\Pi = [\pi_{j,k}] \in \mathbb{R}^{r\times 8}$, along with a learned scale $\gamma_j$ for each basis dimension and a bias $b_k$ for each expert. The router computes an unnormalized score $s_{t,k}$ for expert $k$ on token $t$ as a linear combination of $e_t$’s components:
\begin{equation}
    s_{t,k} = \sum_{j=1}^r \gamma_j\pi_{j,k}e_{t,j} + b_k.
\end{equation}

These scores are passed through a softmax (with temperature $\tau=1$ by default) to obtain routing probabilities $p_{t,k} = \softmax(s_t)k$. Sparse top-1 gating is then applied: for each token $t$, the router selects the single expert with the highest probability and routes $h_t$ to that expert. Only the top-1 expert’s MLP is executed for that token, making the MoE computation efficient and sparse. The expert MLPs are lightweight two-layer feed-forward networks, each with a linear bottleneck architecture and GELU nonlinearity. The output of the chosen expert is merged back into the main branch: the token’s representation is updated as $h_t \leftarrow h_t + \alpha ,\text{Expert} {k^*}(h_t)$, where $\alpha$ is a learned scaling parameter. To ensure the eigenbasis remains stable, we add a small orthonormality loss $L_{\text{ortho}}$ on each router’s $\mathbf{U}$ to penalize deviations from orthogonal, where we define our loss as:
\begin{equation}
    L_{\text{ortho}} = \lambda_{\text{ortho}}| \mathbf{U}^\top \mathbf{U} - I_r|_F^2
\end{equation}
This encourages the basis vectors to stay orthogonal and aligned with variance, preventing collapse of the routing space.

\vspace{-6mm}
\section{Experiments}
\label{sec:exp}
\vspace{-3mm}
To validate the effectiveness of EMoE on large-scale image classification, we train and evaluate our model with ViT backbones on the ImageNet-1K benchmark. We compared our method with three state-of-the-art MoE models, along with the experts' load balance analysis. To ensure a fair comparison, all methods are trained and tested on identical datasets, which include CIFAR, Tiny ImageNet, and ImageNet. Furthermore, we maintain uniformity in the training strategy and hyperparameter settings across all approaches.

\textbf{Datasets.} The CIFAR-10 and CIFAR-100 are obtained from the TensorFlow datasets~\cite{abadi2016tensorflow}. CIFAR-10~\cite{krizhevsky2009learning} consists of 60,000 images, each of size $32\times32$.
CIFAR-100~\cite{krizhevsky2009learning} has the same image size and total count but spans 100 classes, organized into 20 superclasses with both coarse and fine labels.
Tiny ImageNet~\cite{le2015tiny} consists of a dataset of $100,000$ images distributed across 200 classes, with 500 images per class for training, and an additional set of $10,000$ images for testing.
ImageNet-1K~\cite{russakovsky2015imagenet} includes 1{,}281{,}167 training images, 50{,}000 validation images, and 100{,}000 test images over 1{,}000 categories.

\begin{table}[h!]
\vspace{-3mm}
\centering
\caption{Top-1 and top-5 classification accuracy(\%) with different methods on ImageNet}
\label{tab:imagenet_compare}
\begin{tabular}{lccc}
\toprule
\textbf{Method} & \textbf{Top-1 (\%)} &\textbf{Top-5 (\%)} \\
\midrule
V\!-\!MoE   & 87.41 & 97.94 \\
Single-gated MoE   & 72.38 & 93.26 \\
DeepMoE   & 77.12 & 95.07  \\
\midrule
EMoE-ViT-B & 85.32 & 96.45  \\
EMoE-ViT-L & 86.70 & 97.34  \\
EMoE-ViT-H & \textbf{88.14} & \textbf{98.27}  \\
\bottomrule
\end{tabular}
\vspace{-3mm}
\end{table}

\textbf{Baseline models.} In this paper, we choose three leading MoE baselines, which are V-MoE~\cite{riquelme2021scaling}, Single-gated MoE~\cite{royer2023revisiting}, and DeepMoE~\cite{wang2020deep}:
(1) \textbf{V-MoE} replaces a subset of ViT feed-forward layers with sparsely activated expert MLPs and routes image patches (tokens) to a small number of experts. It further introduces Batch Prioritized Routing to adapt per-image compute at test time.
(2) \textbf{Single-gated MoE} revisits a simple, practical design: a lightweight single gate picks one expert per input while a parallel base model branch serves for ensembling and early exit. The training pipeline is asynchronous and aims to avoid router collapse via clustering-based initialization of the gate.
(3) \textbf{DeepMoE} applies MoE ideas within convolutional networks via channel-wise dynamic sparsification, which utilizes a multi-headed sparse gating network to select and rescale channels per input.

\textbf{Results on ImageNet.}
\begin{figure}
    \centering
    \includegraphics[width=0.8\linewidth]{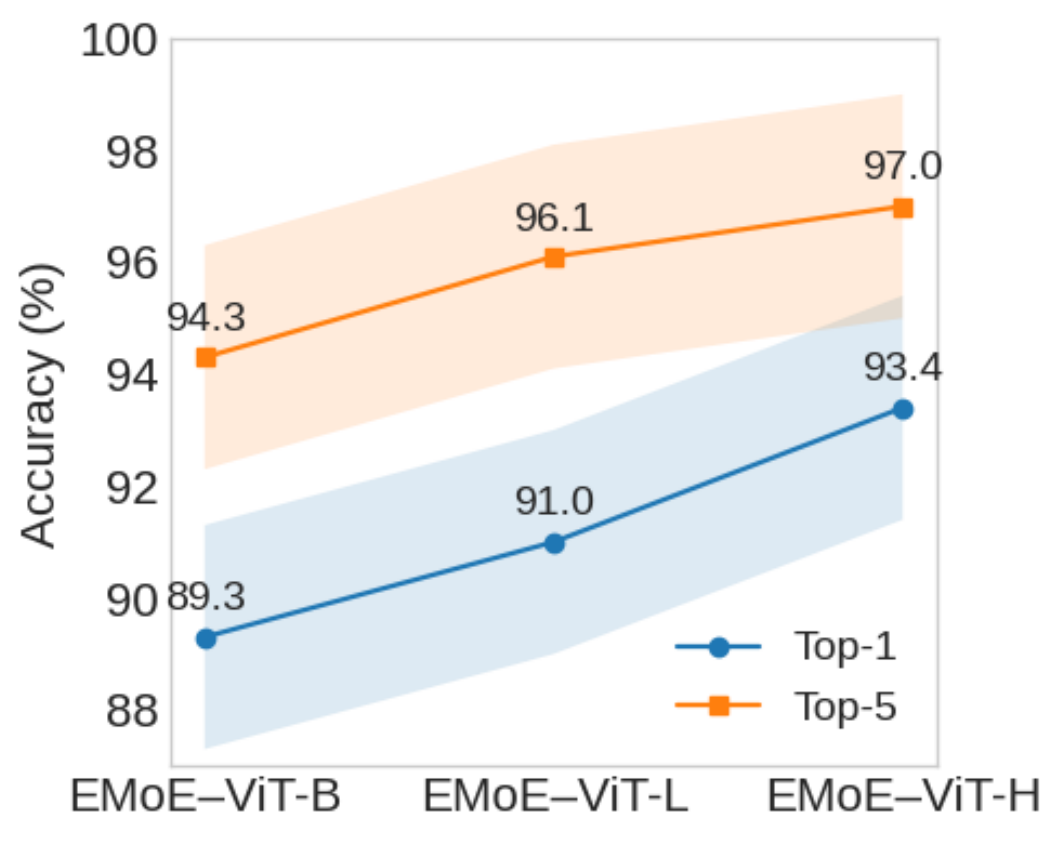}
    \vspace{-5mm}
    \caption{one-shot classification for the proposed model on CIFAR10. We reported $\pm$2\% std.}
    \label{fig:oneshot}
    \vspace{-6mm}
\end{figure}
Our model is trained and evaluated on the challenging ImageNet benchmark. The results, summarized in Table~\ref{tab:imagenet_compare}, demonstrate the competitive performance and superior scalability of the EMoE architecture. Our flagship model, EMoE-ViT-H (EMoE with ViT-H backbone), establishes a new state-of-the-art performance among the evaluated methods. It achieves a Top-1 accuracy of 88.14\% and a Top-5 accuracy of 98.27\%. This result underscores the architectural efficacy of EMoE in large-scale, high-capacity.

Furthermore, the advantages of the EMoE framework are not limited to the largest model. Even the base model, EMoE-ViT-B, with an accuracy of 85.32\%, substantially outperforms both Single-gated MoE and DeepMoE by over \textbf{5\%}. This wide performance gap indicates that the architectural design of EMoE provides a fundamental advantage that is independent of model size and robust across different baselines.

\textbf{Few-shot classification.}
Beyond performance on single dataset, a key measure of a model's utility is the quality of its learned feature representations, which can be effectively evaluated through its ability to generalize from very limited data. The few-shot learning capabilities of EMoE were assessed on CIFAR-10, CIFAR-100, and Tiny-ImageNet.

The one-shot classification performance on CIFAR-10, illustrated in Fig.~\ref{fig:oneshot}, reveals remarkable data efficiency. With only see a single example per class, the EMoE-ViT-H model achieves an impressive Top-1 accuracy of 96.1\% and a Top-5 accuracy of 97.0\%.
More extensive few-shot evaluations using linear probes were conducted on CIFAR-100 and Tiny-ImageNet, with results presented in Table~\ref{tab:fewshot_combined}. In these experiments, EMoE-ViT-H demonstrates a commanding lead over all baseline models across both datasets and shot counts. The performance gap is particularly striking when compared to the strong V-MoE baseline. On the CIFAR-100 10-shot task, EMoE-ViT-H achieves 96.54\% accuracy, a remarkable \textbf{5\%} improvement over V-MoE. This advantage is even more pronounced on the Tiny-ImageNet 10-shot task, where EMoE-ViT-H surpasses V-MoE by almost \textbf{7\%}.

\begin{table}[t]
\centering
\caption{Top-1 (\%) for 5/10-shot linear probes on CIFAR-100 and Tiny-ImageNet-200.}
\label{tab:fewshot_combined}
\begin{adjustbox}{max width=\linewidth,center}
  \setlength{\tabcolsep}{5pt}
  \renewcommand{\arraystretch}{0.95}
  \begin{tabular}{lcccc}
    \toprule
    & \multicolumn{2}{c}{\textbf{CIFAR-100}} & \multicolumn{2}{c}{\textbf{Tiny-ImageNet-200}} \\
    \cmidrule(lr){2-3}\cmidrule(lr){4-5}
    \textbf{Method} & 5-shot & 10-shot & 5-shot & 10-shot \\
    \midrule
    V\!-\!MoE             & 89.49 & 91.26 & 81.12 & 83.16 \\
    Single-gated MoE      & 74.85 & 76.62 & 60.45 & 62.49 \\
    DeepMoE               & 68.81 & 70.58 & 55.46 & 57.49 \\
    \midrule
    EMoE--ViT-B   & 86.92 & 90.00 & 80.03 & 82.81 \\
    EMoE--ViT-L    & 88.83 & 91.37 & 82.05 & 89.97 \\
    EMoE--ViT-H   & \textbf{91.04} & \textbf{96.54} & \textbf{83.71} & \textbf{90.04} \\
    \bottomrule
  \end{tabular}
\end{adjustbox}
\vspace{-5mm}
\end{table}

\begin{figure}
    \centering
    \includegraphics[width=\linewidth]{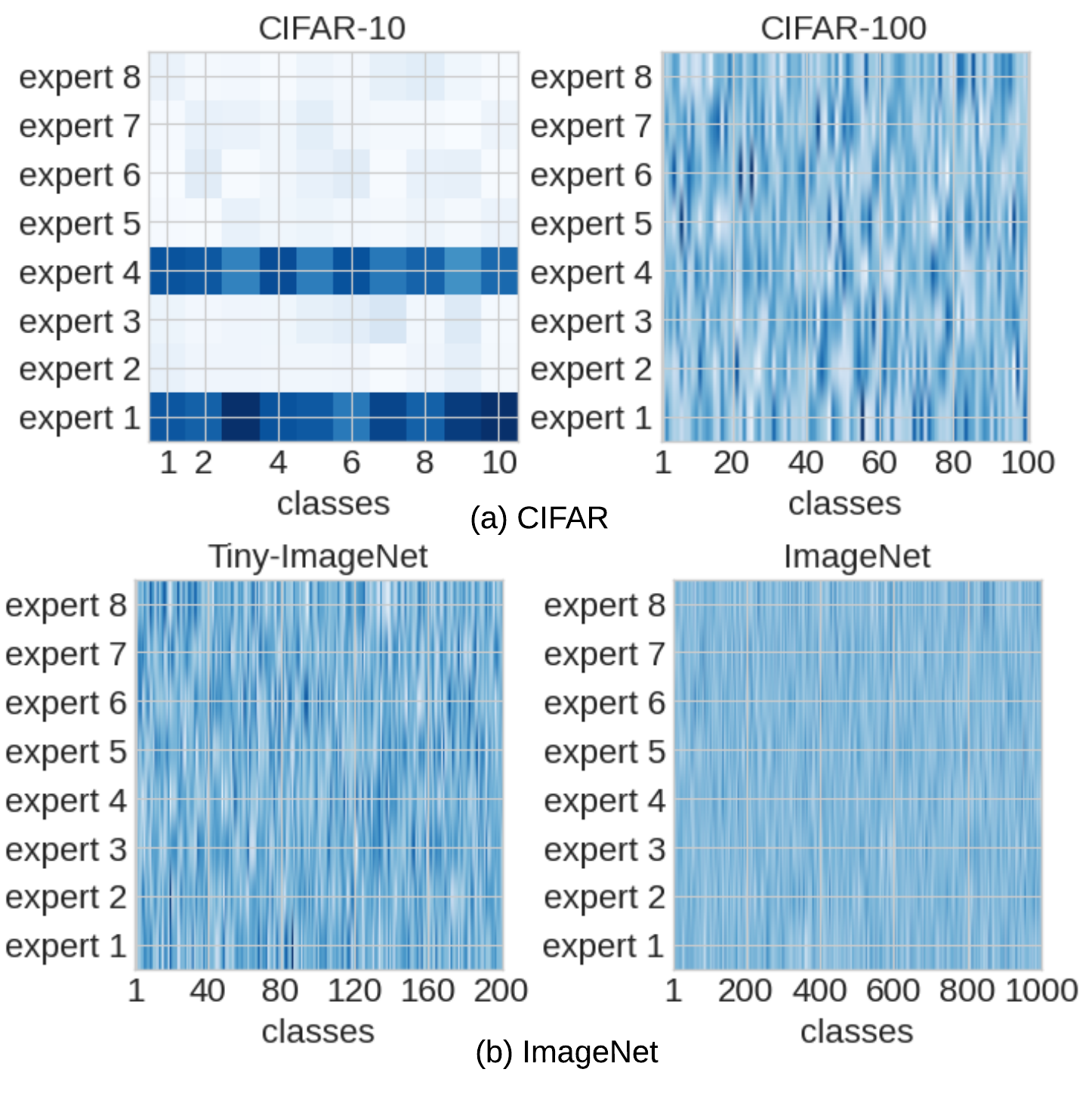}
    \vspace{-8mm}
    \caption{Expert–class routing heatmaps across (a)CIFAR and (b)ImageNet for EMoE. Each panel shows average routed tokens per class (x) and expert (y).}
    \label{fig:load}
    \vspace{-6mm}
\end{figure}

\textbf{Load Balance Analysis.}
To probe how EMoE allocates compute across experts, we visualize the average number of routed tokens per \emph{(expert, class)} in Fig.~\ref{fig:load}. The result reveals that the model's routing strategy is not static but adapts dynamically to the complexity and statistical structure of the dataset. For smaller datasets such as the CIFAR-10 and CIFAR-100 datasets, shown in Fig.~\ref{fig:load}a,  a few experts are preferred for coherent subsets of classes, yet all eight experts remain active, with no rows collapsing to near-zero usage. This indicates that EMoE captures class-level structure without falling into the classic “rich-get-richer’’ dynamic where a handful of experts monopolize traffic. When training on larger datasets, i.e., Tiny-ImageNet and ImageNet, shown in Fig.~\ref{fig:load}b, routing becomes visibly denser and more distributed. Each class is processed by multiple experts and each expert serves a broad, overlapping slice of classes. The resulting pattern is close to uniform at scale, evidencing healthy utilization of \emph{all} experts and robust avoidance of collapse.

This observed adaptability is a core strength of our EMoE architecture. The gating network learns a meaningful and efficient routing policy tailored to the problem's nature. On datasets with clear conceptual separation between classes, it learns to dispatch inputs to distinct specialists. On more complex datasets with high inter-class similarity, it learns a more nuanced, collaborative strategy, effectively forming dynamic ensembles of experts on-the-fly to handle challenging inputs. This effectively prevent peak use for one or several experts with other experts stay vacant.

\textbf{Medical Image Application.} We further evaluate EMoE beyond standard vision benchmarks by applying it to brain age prediction from 3D structural MRI data. This task presents unique challenges due to the high dimensionality and heterogeneity of volumetric medical scans. We adapt EMoE to a 3D CNN backbone, where eigenbasis-guided routing is performed over volumetric patch features. The resulting EMoE model achieves a Mean Absolute Error (MAE) of 2.16 years on brain age estimation, significantly outperforming a standard 3D CNN trained with backpropagation, which yields an MAE of 2.41 years. This corresponds to a \textbf{10.4\%} reduction in prediction error, demonstrating EMoE’s capacity for learning discriminative expert-specialized features in complex biomedical domains.

\vspace{-5mm}
\section{Conclusion}
\label{sec:con}
\vspace{-3mm}
In this work, we introduce EMoE, a sparse expert model that leverages eigenbasis-guided routing and an explicit load-balancing scheme to distribute inputs across specialized experts. This design ensures each expert is effectively utilized and, crucially, mitigates the \emph{rich-get-richer} failure mode observed in conventional MoEs. We demonstrate that EMoE achieves strong performance on ImageNet classification and excels in few-shot learning tasks, highlighting its efficiency and generalization capability. Finally, because routing is aligned with the data’s dominant directions of variation, EMoE dynamically selects specialists while preserving near-uniform expert utilization, which maintains balanced loads without expert monopolies.

Importantly, EMoE’s dynamic expert routing is well-suited to high-dimensional, heterogeneous data typical in biomedical applications. For example, our experiments shows that in brain age prediction, EMoE can adaptively route different brain regions or imaging modalities to dedicated experts, capturing complex anatomical patterns while maintaining balanced expert usage. These capabilities position EMoE as a promising approach for biomedical applications, and future work will evaluate its performance on these challenges.
\bibliographystyle{IEEEbib}
\bibliography{refs}

\end{document}